\documentclass{article} % For LaTeX2e
\usepackage{iclr2027_conference,times}

\usepackage{graphicx}
\usepackage{booktabs}
\usepackage{subcaption}
\usepackage{multirow}
\usepackage{rotating}

\usepackage[most]{tcolorbox}
\newcommand{\titletext}{How To Train Your World Model: Fine-tuning vs RAG for LM-based World Modeling}
\usepackage{amsmath,amsfonts,bm}

\def\eqref#1{equation~\ref{#1}}
\def\1{\bm{1}}

\DeclareMathAlphabet{\mathsfit}{\encodingdefault}{\sfdefault}{m}{sl}
\SetMathAlphabet{\mathsfit}{bold}{\encodingdefault}{\sfdefault}{bx}{n}

\usepackage{hyperref}
\usepackage{url}
\usepackage[normalem]{ulem}

\title{\titletext}

\author{Dhananjay Ashok\thanks{Work done during internship at CapitalOne} \\
Information Sciences Institute \\
University of Southern California \\
\texttt{ashokd@usc.edu} \\
\And
Shantanu Agarwal \\
Capital One \\
\texttt{shantanu.agarwal1@capitalone.com} \\
\And
Vivek Datla \\
Capital One \\
\texttt{vivek.datla@capitalone.com} \\
\AND
Jonathan May \\
Information Sciences Institute \\
University of Southern California \\
\texttt{jonmay@isi.edu} \\
\And
Alfy Samuel \\
Capital One \\
\texttt{alfy.samuel@capitalone.com}
}

\iclrfinalcopy % Uncomment for camera-ready version, but NOT for submission.
\begin{document}

\maketitle

\begin{abstract}
World models (WMs) simulate the transition dynamics of environments, enabling agents to plan over the consequences of their actions. In text-based environments, fine-tuning a Language Model (LM) to serve as a WM has emerged as a dominant paradigm. 
However, despite the widespread success of non-parametric approaches such as Retrieval Augmented Generation (RAG), retrieval over past experience for LM-based world modelling remains underexplored. We conduct a systematic evaluation across five diverse environments spanning embodied, web navigation and social settings, comparing fine-tuning and RAG-based approaches for LM-based world modelling. 
Our study reveals that fine-tuning often outperforms RAG, with fine-tuned WMs enabling agents to obtain higher rewards on 15/20 settings. 
However, there are several cases where RAG proves superior, and there are often cases where simple prompting baselines match or outperform finetuning. 
While both construction paradigms benefit from additional and more diverse exploration, RAG-based approaches prove more data-efficient, while fine-tuning approaches disproportionately benefit from scaling the amount of experience collected. 
With a focus on RAG-based WMs, we devise a procedure that uses counterfactual intervention to estimate the error rate of the retrieval stage, and show that retrievers consistently surface suboptimal transitions from the experience buffer. 
Hoping to address this failing, we study a variety of query reformulation strategies, demonstrating that a hierarchical approach outperforms the traditional retrieval pipeline. 
Finally, we compose our findings into a hybrid world modelling system that parametrically captures core environment dynamics, while learning to rely on retrieval from an actively maintained memory store. 
Our hybrid system consistently outperforms other methods across multiple environments and models, showcasing the robustness of the approach and the applicability of our findings.
\end{abstract}

\section{Introduction}
\label{sec:introduction}

Agentic systems increasingly improve by scaling experience within complex environments, shifting away from a reliance on dense human annotation and engineered reward systems~\citep{silver2025welcome}. A dominant approach to scaling experience is the development of \textit{world models} (WMs)—internal simulators that learn the underlying transition dynamics of an environment~\citep{ha2018world}. 
Using a WM, agents can go beyond simply reacting to the current state and instead plan by considering the consequences of their actions before executing them. 

To leverage the rich knowledge priors encoded in Language Models (LMs)~\citep{petroni2019language}, recent work has explored using LMs as the backbone for world models~\citep{gu2024your}. These methods follow the traditional approach to world modelling of collecting exploratory trajectories on a training set, and then finetuning a model to predict future observations~\citep{watter2015embed, chae2025web}. When applied to LMs, this parametric approach yields WMs that can boost agent performance on a variety of environments, from embodied settings to web-domains~\citep{li2026word}.

Particularly prominent in the LM domain is the use of non-parametric approaches, such as retrieval-augmented generation (RAG)~\citep{arslan2024survey}.
Relying on the in-context capabilities of LMs, RAG make no changes to the weights, but rather provides examples and additional information relevant to the task at hand to help ground the LM's response. While both fine-tuning and RAG have their individual strengths~\citep{balaguer2024rag}, direct comparisons demonstrate that RAG-based approaches can often substantially outperform parametric adaptation in knowledge-acquisition tasks~\citep{ovadia2024fine, soudani2024fine}. However, despite the power of non-parametric approaches, their application to LM-based world modelling remains underexplored. 

In this work, we systematically compare strategies for constructing LM-based WMs across five environments, including embodied tasks (ALFWorld~\citep{shridhar2020alfworld}, TextWorld~\citep{cote2018textworld}), web navigation (WebShop~\citep{yao2022webshop}), and social conversations (Sotopia~\citep{zhou2024sotopia}, Chameleon~\citep{karabag2025llms}). We construct both fine-tuned and RAG-based WMs and compare their performance on the axes of WM fidelity, as well as agent utility (Section~\ref{sec:metrics}). %Specifically, we ask the following research questions:

%\textcolor{blue}{DJ: I'm not a fan of research questions. I think they take too much space for the clarity they add.}
%\begin{itemize}
%    \item \textbf{RQ1}: How does fine-tuning compare against RAG for LM-based world modeling?
%    \item \textbf{RQ2}: How does the composition and size of the replay buffer affect each method?
%    \item \textbf{RQ3}: How accurate is retrieval over experienced trajectories?
%    \item \textbf{RQ4}: Can we combine fine-tuning and RAG to make a superior system?
%\end{itemize}

Prior studies comparing RAG and fine-tuning frequently find RAG to be dominant in tasks such as knowledge-grounded question answering~\citep{ovadia2024fine}. Our results show the opposite: RAG underperforms fine-tuning on 15/20 model-environment combinations, but maintains a strong lead on specific environments.  Additionally, on 9/20 settings, either fine-tuning or RAG is outperformed by a simple prompting baseline, demonstrating that the ideal approach is highly dependent on the specific environment in question.

Ablations over the buffers used to construct the WMs reveal approach-specific trends. Both RAG and fine-tuning benefit from larger and more diverse exploration data; however, RAG is more data-efficient, while fine-tuning exhibits more persistent returns to increases in scale.

With a focus on the yet unexplored RAG-based WM approach, we devise a counterfactual intervention experiment to estimate the error rate of the retrieval stage. Our findings are stark and clear: the retrieval step fails consistently with RAG-based WMs, with recall@5 falling below 0.7. Motivated by this finding, we investigate more complex ways of executing retrieval over observations and actions. We find that a hierarchical retrieval paradigm, which selects based on observation similarity before re-ranking based on action similarity, outperforms all other approaches. 

Finally, we introduce a hybrid paradigm which combines the parametric and RAG-based approaches to construct a world model. Our hybrid approach outperforms every other method on 17/20 model-environment combinations, approaching 77\% of the utility of an oracle upper bound (compared to 60\% by the next best method). On 45\% of environment-model combinations, our hybrid exceeds 90\% of the performance of the oracle model, establishing the power of the method. 

\begin{figure}
    \centering
    \includegraphics[width=\linewidth]{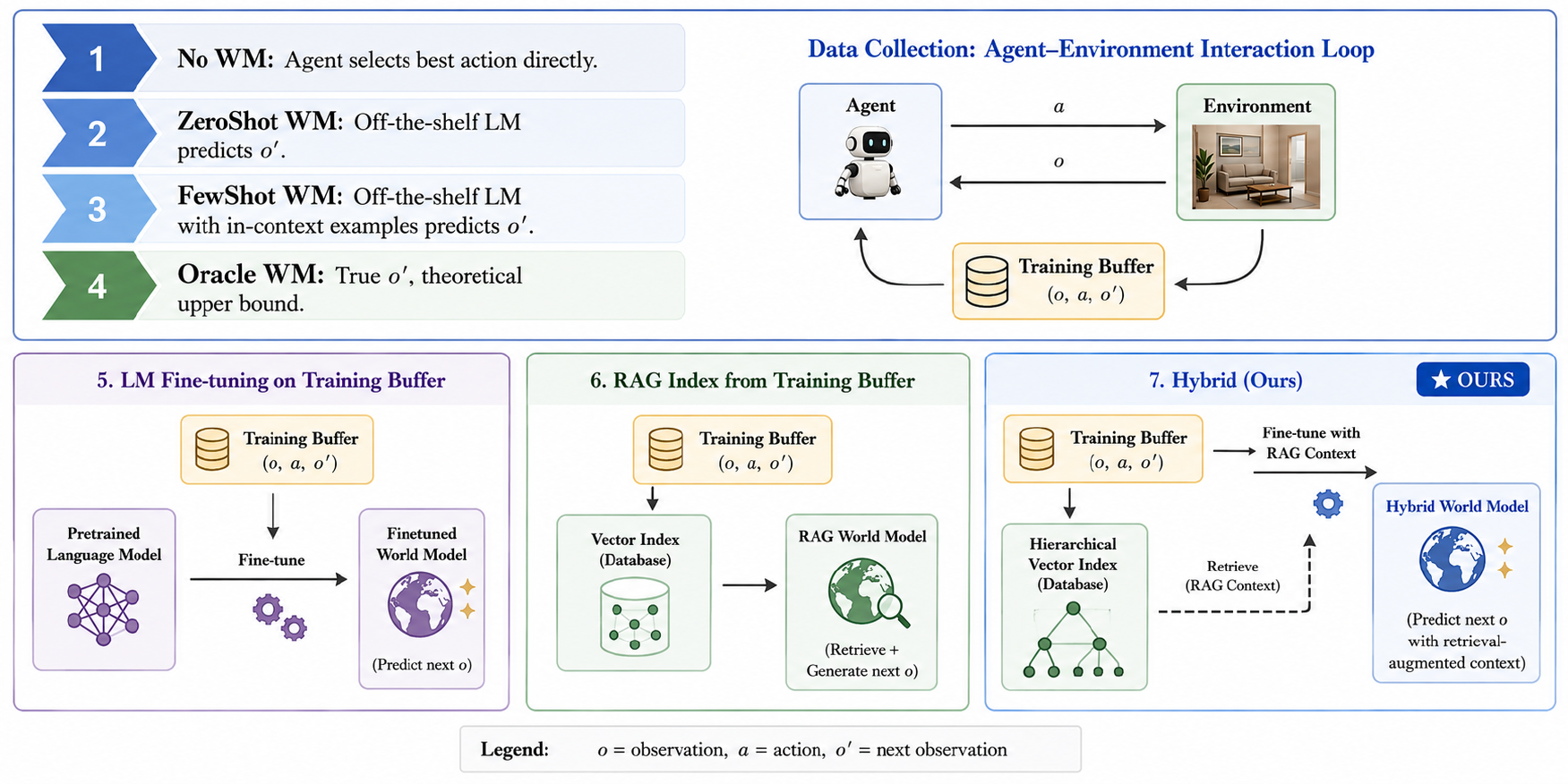}
    \caption{We investigate the performance of fine-tuning vs RAG for the task of LM-based world modeling. Finally, we distill our insights into a hybrid methodology that combines fine-tuning with a hierarchical index design to robustly improve the performance of LM-based WMs.}
    \label{fig:study}
\end{figure}

\section{Background and Related Work}
\label{sec:related-work}

We tackle the generic problem of an agent in an environment, formally represented as a partially observable Markov decision process (POMDP)~\citep{puterman1990markov}, defined by the 7-tuple:
\begin{equation}
    \mathcal{M} = \langle \mathcal{S}, \mathcal{A}, \mathcal{P}, \mathcal{R}, \Omega, \mathcal{O}, \gamma \rangle
\end{equation}
where $\mathcal{S}$ is the state space, $\mathcal{A}$ is the action space, and $\Omega$ is the observation space. The transition probability function $\mathcal{P}(s' \mid s, a)$ denotes the probability of transitioning to state $s' \in \mathcal{S}$ given the current state $s \in \mathcal{S}$ and action $a \in \mathcal{A}$. At each time step, the agent receives an observation $o \in \Omega$ determined by the observation function $\mathcal{O}(o \mid s', a)$, alongside a scalar reward from the reward function $\mathcal{R}(s, a)$. 
Finally, $\gamma \in [0, 1)$ represents the discount factor. Given a POMDP, the goal of an agent is to select actions at each step such that the expected total reward is maximized:
\begin{equation}
    \pi^* = \arg \max_{\pi} \mathbb{E}_{\tau \sim \pi} \left[ \sum_{t=0}^{\infty} \gamma^t \mathcal{R}(s_t, a_t) \right]
\end{equation}
where $\tau = (o_0, a_0, \ldots a_{t-1}, o_t)$  denotes a trajectory sampled from the environment under policy $\pi$.

\subsection{World Models}
World Models estimate the transition function $\mathcal{P}(s' \mid s, a)$~\citep{ha2018world}. However, since agents are only exposed to observations, world models typically learn a transition function based on observations, $\mathcal{P}(o' \mid o, a)$. World models have a variety of uses, including the replacement of expensive environment calls during agent learning~\citep{hafner2020mastering}, and the enabling of planning over the potential consequences of an agent's selected actions~\citep{sekar2020planning}. 

The traditional approach to training a world model has its roots in image generation~\citep{hafner2019dream}. First, in the exploration phase, agents collect an experience replay buffer on training splits of the environment $\mathcal{D} = \left\{ \tau^{(i)} \right\}_{i=1}^N$. We then use the transitions from the buffer to train a world model: $\text{WM}: \mathcal{O} \times \mathcal{A} \to \mathcal{O}$ by minimizing the prediction error: $\mathcal{L}(\theta) = \mathbb{E}_{\tau \sim \mathcal{D}} \left[ \text{loss}(\text{WM}_\theta(o_t, a_t), o_{t+1}) \right]$.

Since their emergence as powerful aggregators of human knowledge and reasoning skills, there has been increased interest in using LMs to serve as the foundations for WMs~\citep{gu2024your, chae2025web}. These works follow the traditional paradigm, and fine-tune a LM to predict the next observation with a next-token-prediction loss.  

A strong alternative to finetuning is non-parametric adaptation or retrieval-augmented generation (RAG), often used to facilitate memory in agentic systems~\citep{kim2025coex, ferraz2026retrieval}. Work on directly using RAG for world modeling has been limited to initial explorations. Prior work~\citep{mei2025r} creates WMs with RAG mechanisms over external, tutorial data to directly reduce the degradation of long-horizon reasoning. However, this paradigm requires access to manually curated external datasets, a resource which is hard to gather for many complex environments. To the best of our knowledge, the only work that uses RAG over self collected experience replay buffers to construct LM-based WMs are ~\citet{yang2025efficient} and ~\citet{zhang2026self}, however they do not compare against fine-tuning, leaving the comparative strength of the approach unproven. 

\subsection{Systematic Comparisons of LM-based World Models}
Prior work in domains such as knowledge acquisition has often directly compared fine-tuning and RAG-based approaches. While neither method is strictly superior to the other~\citep{balaguer2024rag}, there is widespread agreement that RAG-based approaches enable more effective knowledge acquisition, especially for less popular or domain-specific knowledge~\citep{ovadia2024fine, soudani2024fine, pingua2025medical}. Existing comparisons of LM-based WMs such as \citet{li2026word}, establish a comprehensive evaluation framework for fine-tuned LM-based world models, demonstrating that fine-tuning substantially enhances a LM's next-state prediction fidelity and long-horizon consistency. However, none of these works consider any RAG-based WMs in their comparisons, leaving open the question of how fine-tuning performs relative to RAG in this domain. Our work addresses this gap and conducts the first systematic comparison of fine-tuning and RAG-based approaches for the task of LM-based world modeling.

\section{Fine-tuning v.s. RAG for World Modeling}
\label{sec:experiment}

To facilitate a comprehensive evaluation, we compile a diverse testbed of five text-based environments spanning the domains of embodied navigation, web navigation, and social conversations.

\noindent\textbf{Embodied Navigation:} These environments benchmark embodied decision-making within text-simulated physical spaces. Specifically, we use the ALFWorld and TextWorld environments. In ALFWorld, agents complete household tasks (e.g., ``put a clean apple in the fridge'') by navigating rooms and interacting with objects. TextWorld similarly offers procedurally generated text-based quests, such as gathering ingredients to cook a specific recipe. Agents receive a sparse binary reward ($+1$) upon the successful completion of the high-level task. In these tasks, the WM must learn to predict updated room descriptions, inventory state, or interaction outcomes.

\noindent\textbf{Web Navigation:} We use the WebShop environment, a simulated e-commerce environment where agents must navigate a web interface to purchase specific items based on natural language instructions (e.g., ``buy a red men's running shoe size 10 under \$50''). The agent interacts by issuing search queries and clicking text-based buttons. The agent receives a dense scalar reward between 0 and 1 based on the attribute and semantic overlap between the purchased item and the target instruction. The WM must learn to predict search results pages or product details pages.

\noindent\textbf{Adversarial Negotiation:} To evaluate the agent's ability to achieve complex social objectives, we simulate two-agent, mixed-motive conversations using scenarios adapted from SOTOPIA. The agent must negotiate effectively against a simulated partner (powered by an LM) endowed with hidden information and competing priorities. For example, the agent may be tasked with securing a \$40 loan from a friend who is secretly saving for a major purchase. The agent receives a scalar reward proportional to the resource concessions obtained from the conversation partner by the end of the dialogue. Given the conversation transcript history and the agent's proposed next utterance, the WM aims to predict the simulated partner's subsequent conversational response.

\noindent\textbf{Deceptive Communication:} This environment tests the agent's capacity for social reasoning and managing implicit information flow via the social deduction game ``Chameleon.'' In this game, players receive a secret word, except for one player (the Chameleon). Players take turns stating a single related word to prove they know the secret without revealing it to the Chameleon. We simulate a 3-player game where the agent always acts first and is a standard player (non-Chameleon). This reduces to a single-turn decision problem: selecting a word that balances signalling knowledge to peers while obfuscating it from the Chameleon. After every player has stated a word, an external judge guesses which one was the Chameleon. The agent is rewarded if the Chameleon is correctly accused by the judge, is given zero reward if the other (non-Chameleon) player is accused, and a negative reward if it is accused. Since this is a single-step environment, the WM is trained to predict the subsequent words generated by the other two players, as well as the judge's verdict, conditioned on the agent's chosen word.

For implementation details, refer to Appendix~\ref{app:env-details}. 

\subsection{Methods}

We evaluate a range of baseline LM world models. These methods do not require active construction and serve as reference points for the performance of fine-tuned or RAG-based WMs:

\noindent \textbf{No WM (No):} This setting acts as our foundational, model-free baseline where the agent operates entirely without a world model. Instead of rolling out simulated future states, the agent's policy directly selects an action based purely on the current observation $o_t$.

\noindent \textbf{Zero-Shot WM (ZS):} We evaluate the inherent, out-of-the-box reasoning capabilities of pre-trained LMs. In this setting, the LM is prompted with the current observation $o_t$ and action $a_t$, and is tasked with predicting the subsequent state $o_{t+1}$ relying solely on its pre-trained parametric knowledge.

\noindent \textbf{Few-Shot WM (FS):} To account for the in-context learning capabilities of modern LMs~\citep{brown2020language}, we augment the prompt with $k=5$ randomly sampled environmental transitions $(o_i, a_i, o_{i+1})$. This evaluates whether standard few-shot demonstration is sufficient to accurately model environment dynamics without the need for targeted retrieval mechanisms. 

\noindent \textbf{Ground-Truth WM (Oracle):} We query the actual underlying environment to perfectly resolve the next state. While unachievable in practical deployment, this oracle serves as a strict theoretical upper bound, isolating the downstream agent's performance given a flawless transition model.

The other methods require an active exploration phase where the agent collects an experience replay buffer on the training split of the environment $\mathcal{D} = \left\{ \tau^{(i)} \right\}_{i=1}^N$. Given this buffer, we construct:

\noindent\textbf{Fine-tuned WM (FT):} We fine-tune the LM on verbalised transitions from the experience replay buffer. Specifically, we optimise the loss: 

\begin{equation}
    \mathcal{L}_{\text{NTP}}(\theta) = \mathbb{E}_{\tau \sim \mathcal{D}} \left[ -\sum_{k=1}^{K} \log p_\theta \left( z^k_{t+1} \mid c_t, z^{<k}_{t+1} \right) \right]
\end{equation}
where $c_t = \text{tokenize}(o_t, a_t)$ represents the serialised token sequence of the observation and current action, and $z_{t+1} = (z^1_{t+1}, \dots, z^K_{t+1}) = \text {tokenise} (o_{t+1})$ denotes the sequence of $K$ discrete tokens comprising the target next observation. The LM models the probability $p_\theta$ of each subsequent token $z^k_{t+1}$ conditioned on the context prefix $c_t$ and all preceding observation tokens $z^{<k}_{t+1}$.

\noindent\textbf{Retrieval Augmented Generation WM (RAG):} We construct an index over the transitions in the replay buffer by representing each transition as a key string: $o\circ a$, where $\circ$ is string concatenation. Each key string is embedded in a vector space with Nemotron-Embed-8B~\citep{xu2025omni}. During inference, given an observation $o_t$ and candidate action $a_t$, we compute the embedding of $o_t\circ a_t$, retrieve the $k=5$ most similar transitions \{$(o_i, a_i, o_{i+1}), \ldots$\} from the experience replay buffer and serve them to the LM in-context as we request a prediction on the next observation. 

\subsection{World Model Deployment and Evaluation}
\label{sec:metrics}
We choose a consistent pipeline of deployment for all WMs. All of our experiments involve a frozen, off-the-shelf LM as a policy model/agent (in our study we use Llama3-8B~\citep{grattafiori2024llama}, Mistral-7B~\citep{Jiang2023Mistral7}, Gemma-4-26B-A4B~\citep{team2025gemma} and Gemma-4-31B) which uses a world model to plan over its next action. Specifically, the agent sees observation $o_t$ and instead of predicting a single action $a_t$, predicts a list of candidate actions $a_{t, 1}, \ldots a_{t, m}$. The world model under evaluation is used to predict the next observation for each of these candidate actions: $\tilde{o}_1, \ldots \tilde{o}_m$. We then use the policy-agent LM to estimate the value of the predicted observation on a scale of 1-10. The action which led to the most highly rated next observation is selected. In practice, we set $m=5$. For exact prompts, hyperparameters and implementation details, see Appendix~\ref{app:wm-deployment}. We follow prior work~\citep{li2026word} and evaluate each WM on two axes: 

\noindent\textbf{Fidelity:} 
World models predict the next observation. A high-fidelity world model is one that more accurately predicts the next observation. Given that observations can be represented as strings, we use multiple measures of string similarity. Our metrics include lexical overlap (BLEU), semantic overlap (BertScore) and general meaning (Gemma-4-31B-as-a-judge on a scale of 1-10). 
However, initial results suggested that both BLEU and BertScore were problematic metrics, exhibiting little variance over runs with drastically different performance/behavior. Hence, we only report LM-as-a-judge in the main text. See Appendix~\ref{sec:appcomplete} for complete results. 

\noindent\textbf{Utility:} The eventual goal of world modelling is to create agentic systems which achieve higher reward. Hence, we additionally evaluate WMs by the total reward that an agent obtains when using it for planning. 
A WM with higher utility, by definition, leads to a higher boost in agent performance. 
To enable comparisons across environments, we scale all total rewards by the total reward obtained by an agent that has access to the oracle world model. 

We compute 95\% confidence intervals using a paired bootstrapped t-test~\citep{koehn-2004-statistical}. 

% Please add the following required packages to your document preamble:
% \usepackage{booktabs}
% \usepackage[table,xcdraw]{xcolor}
% Beamer presentation requires \usepackage{colortbl} instead of \usepackage[table,xcdraw]{xcolor}
\begin{table}[]
\centering
\footnotesize
\setlength{\tabcolsep}{4pt}
\begin{tabular}{@{}lllllllllll@{}}
\toprule
Environment & Model & No & \multicolumn{2}{c}{ZS} & \multicolumn{2}{c}{FS} & \multicolumn{2}{c}{FT} & \multicolumn{2}{c}{RAG} \\ \midrule
 & & Util & Util & Fid & Util & Fid & Util & Fid & Util & Fid \\ \midrule
ALF          & Llama    & $0.36$         & $\underline{0.43}$ & $\underline{4.15}$ & $0.25$         & $3.96$         & $\mathbf{0.46}$ & $\mathbf{4.78}$ & $0.36$         & $4.09$         \\
             & Mistral  & $0.00$         & $0.30$         & $3.95$         & $0.20$         & $3.82$         & $\mathbf{0.54}$ & $\mathbf{4.65}$ & $\underline{0.39}$ & $\underline{4.12}$ \\
             & Gem-26B  & $0.79$         & $0.80$         & $4.33$         & $0.76$         & $4.16$         & $\mathbf{0.86}$ & $\mathbf{5.10}$ & $\underline{0.81}$ & $\underline{4.87}$ \\
             & Gem-31B  & $0.82$         & $0.67$         & $4.23$         & $0.72$         & $5.01$         & $\mathbf{0.89}$ & $\mathbf{6.18}$ & $\underline{0.86}$ & $\underline{5.11}$ \\ \midrule
Text         & Llama    & $0.00$         & $\mathbf{0.16}$ & $3.40$         & $0.06$         & $2.87$         & $\underline{0.15}$ & $\underline{4.04}$ & $0.11$         & $\mathbf{4.15}$ \\
             & Mistral  & $0.49$         & $\mathbf{0.63}$ & $3.45$         & $0.50$         & $2.90$         & $\underline{0.57}$ & $\underline{4.10}$ & $0.55$         & $\mathbf{4.20}$ \\
             & Gem-26B  & $0.00$         & $\mathbf{0.22}$ & $3.61$         & $0.14$         & $3.13$         & $\underline{0.19}$ & $\underline{4.28}$ & $0.17$         & $\mathbf{4.36}$ \\
             & Gem-31B  & $0.85$         & $0.76$         & $5.20$         & $0.63$         & $4.82$         & $\mathbf{0.92}$ & $\mathbf{5.86}$ & $\underline{0.90}$ & $\underline{5.63}$ \\ \midrule
Web          & Llama    & $0.04$         & $0.07$         & $4.77$         & $0.09$         & $3.78$         & $\mathbf{0.49}$ & $\mathbf{6.06}$ & $\underline{0.24}$ & $\underline{5.08}$ \\
             & Mistral  & $0.00$         & $0.17$         & $4.65$         & $0.25$         & $3.85$         & $\mathbf{0.58}$ & $\mathbf{6.15}$ & $\underline{0.33}$ & $\underline{5.15}$ \\
             & Gem-26B  & $0.08$         & $0.12$         & $4.81$         & $0.17$         & $4.11$         & $\mathbf{0.53}$ & $\mathbf{6.27}$ & $\underline{0.35}$ & $\underline{5.34}$ \\
             & Gem-31B  & $0.90$         & $0.60$         & $4.01$         & $0.69$         & $4.29$         & $\mathbf{0.94}$ & $\mathbf{6.43}$ & $\underline{0.92}$ & $\underline{5.81}$ \\ \midrule
Cha          & Llama    & $\underline{0.17}$ & $-0.50$        & $2.56$         & $-0.10$        & $2.70$         & $-0.52$        & $\underline{3.54}$ & $\mathbf{0.33}$ & $\mathbf{3.87}$ \\
             & Mistral  & $-4.50$        & $-3.74$        & $2.45$         & $\underline{-3.07}$ & $2.60$         & $-3.88$        & $\underline{3.45}$ & $\mathbf{0.64}$ & $\mathbf{3.95}$ \\
             & Gem-26B  & $\underline{0.50}$ & $0.07$         & $3.22$         & $0.48$         & $3.42$         & $0.48$         & $\underline{3.70}$ & $\mathbf{0.66}$ & $\mathbf{4.31}$ \\
             & Gem-31B  & $\underline{0.78}$ & $-0.05$        & $3.28$         & $0.41$         & $\underline{4.16}$ & $0.73$         & $3.83$         & $\mathbf{0.87}$ & $\mathbf{5.17}$ \\ \midrule
Soto         & Llama    & $0.75$         & $\underline{0.89}$ & $4.50$         & $0.82$         & $4.19$         & $\mathbf{0.90}$ & $\underline{4.61}$ & $0.86$         & $\mathbf{4.77}$ \\
             & Mistral  & $0.95$         & $\underline{0.96}$ & $4.40$         & $\underline{0.96}$ & $4.24$         & $\mathbf{0.97}$ & $\underline{4.58}$ & $\mathbf{0.97}$ & $\mathbf{4.86}$ \\
             & Gem-26B  & $0.85$         & $0.87$         & $4.61$         & $0.87$         & $4.42$         & $\mathbf{0.90}$ & $\underline{4.92}$ & $\underline{0.89}$ & $\mathbf{5.12}$ \\
             & Gem-31B  & $0.96$         & $0.96$         & $4.73$         & $0.64$         & $4.62$         & $\mathbf{0.98}$ & $\underline{4.81}$ & $\underline{0.97}$ & $\mathbf{4.87}$ \\ \bottomrule
\end{tabular}
\caption{Utility and Fidelity of LM-based WMs, with best in \textbf{bold} and second best \underline{underlined}. Fine-tuning is the superior standalone method, achieving superior utility on 15/20 settings. RAG maintains a consistent advantage over the baseline approaches and outperforms fine-tuning on the Chameleon environment. Results are significant at $p<0.05$, see Appendix~\ref{sec:appcomplete} for full table.}
\label{table:core}
\end{table}

\subsection{Results}
When evaluating end-to-end task success, fine-tuning is the superior standalone method, achieving superior performance on 15/20 settings. RAG has its strengths; it maintains a consistent advantage over the baseline approaches and outperforms fine-tuning in the Chameleon environment. 

However, despite the superiority of actively constructed world models, they are not universally optimal. In 9/20 evaluated settings, either FT or RAG is outperformed by non-constructed baselines (No WM, ZS, or FS), reinforcing that the ideal approach remains highly environment-dependent.

Interestingly, when isolating the textual accuracy of the predicted next observations, the trend becomes less clear. While FT (4.87) slightly outpaces RAG (4.74) in raw generative accuracy, we notice that in environments such as TextWorld and Sotopia, RAG achieves greater average Fidelity than fine-tuned WMs, despite providing less Utility. This discrepancy correlates with findings from other studies on WMs~\citep{li2026word}, which suggest that fidelity, while loosely correlated with total reward, can imply a different ranking over methods.

\section{Ablations Over Experience Replay Collection}
\label{sec:ablations}

Since both fine-tuning and RAG-based WMs use replay buffers in their construction, we ask how the nature of the buffer affects the performance of each method. We vary the buffer over two axes:

\noindent\textbf{Scale:} We subsample the number of points in the buffer to evaluate how performance varies with data scale. Specifically, we construct WMs with experience replay buffers of $n$ transitions with $n\in \{10, 20, 40, 80, 160, 320, 640, 1280, 2560\}$.

\noindent\textbf{Exploration:} Previous experiments constructed the buffers using an agent that greedily attempted to solve a task in the training environment. An alternate strategy is to mix exploitation and exploration and have the agent take unguided, exploratory actions in the environment $\epsilon\%$ of the time. This approach can potentially increase the diversity of states encountered during exploration, enhancing the WM's effective coverage. In practice, we implement the exploratory action selection mechanism with specialised prompts (see Appendix~\ref{app:exploration}) and set $\epsilon\in\{0, 25, 50, 75, 100\}$.

\subsection{Results:}
We observe distinct scaling behaviours and exploration preferences between fine-tuning (FT) and retrieval-augmented generation (RAG) when varying the experience replay buffer.

\noindent\textbf{Buffer Scale:} Varying the size of the replay buffer reveals contrasting scaling trends. Across TextWorld and Sotopia, RAG achieves strong, near-peak performance with as few as $n=40$ transitions. However, it tends to plateau or slightly degrade at higher scales. Fine-tuning fails at low scales, yielding near-zero rewards. It consistently reaches an inflexion point around $n=80$, where performance rapidly accelerates and begins scaling predictably with more data.

\noindent\textbf{Exploration Ratio:} Injecting exploratory actions during buffer collection improves downstream world model performance, but the optimal distribution diverges by method. For both FT and RAG, completely unguided exploration ($\epsilon=100\%$) and pure exploitation ($\epsilon=0\%$) yield the lowest total rewards across all four environments. Fine-tuning universally peaks at $\epsilon=75\%$ (e.g., reaching a total reward of 15.04 in TextWorld and 0.88 in Sotopia). The parametric model relies on encountering a vast, diverse set of edge-case states during training to accurately internalise the environment's transition dynamics. Conversely, RAG universally peaks at a moderate $\epsilon=50\%$. We hypothesise that because RAG relies on similarity matching in a vector space, excessive random exploration introduces disconnected, noisy transitions that dilute the relevance of the retrieved context.

\begin{figure}[htbp]
    \centering
    \includegraphics[width=0.49\linewidth]{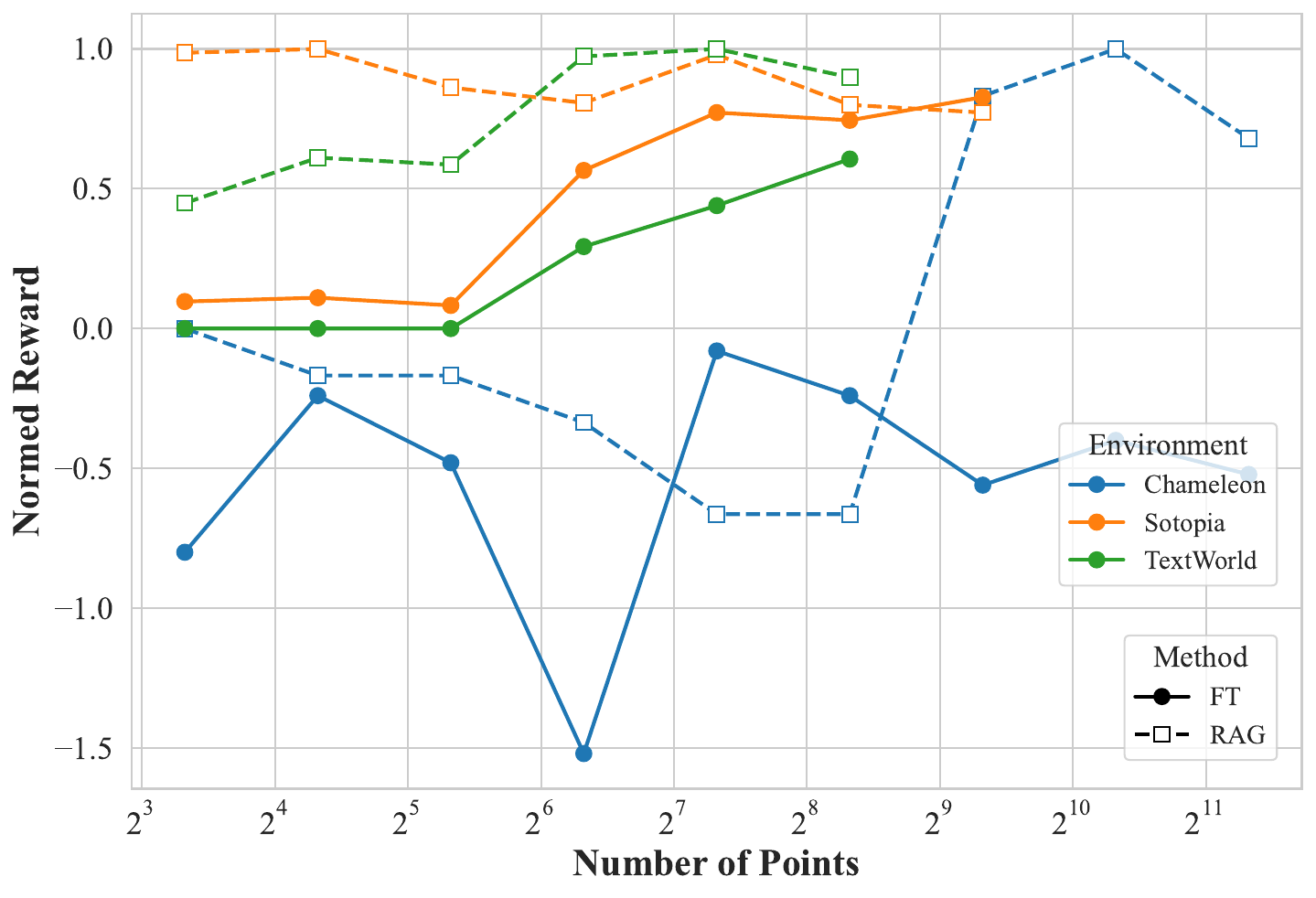}
    \includegraphics[width=0.49\linewidth]{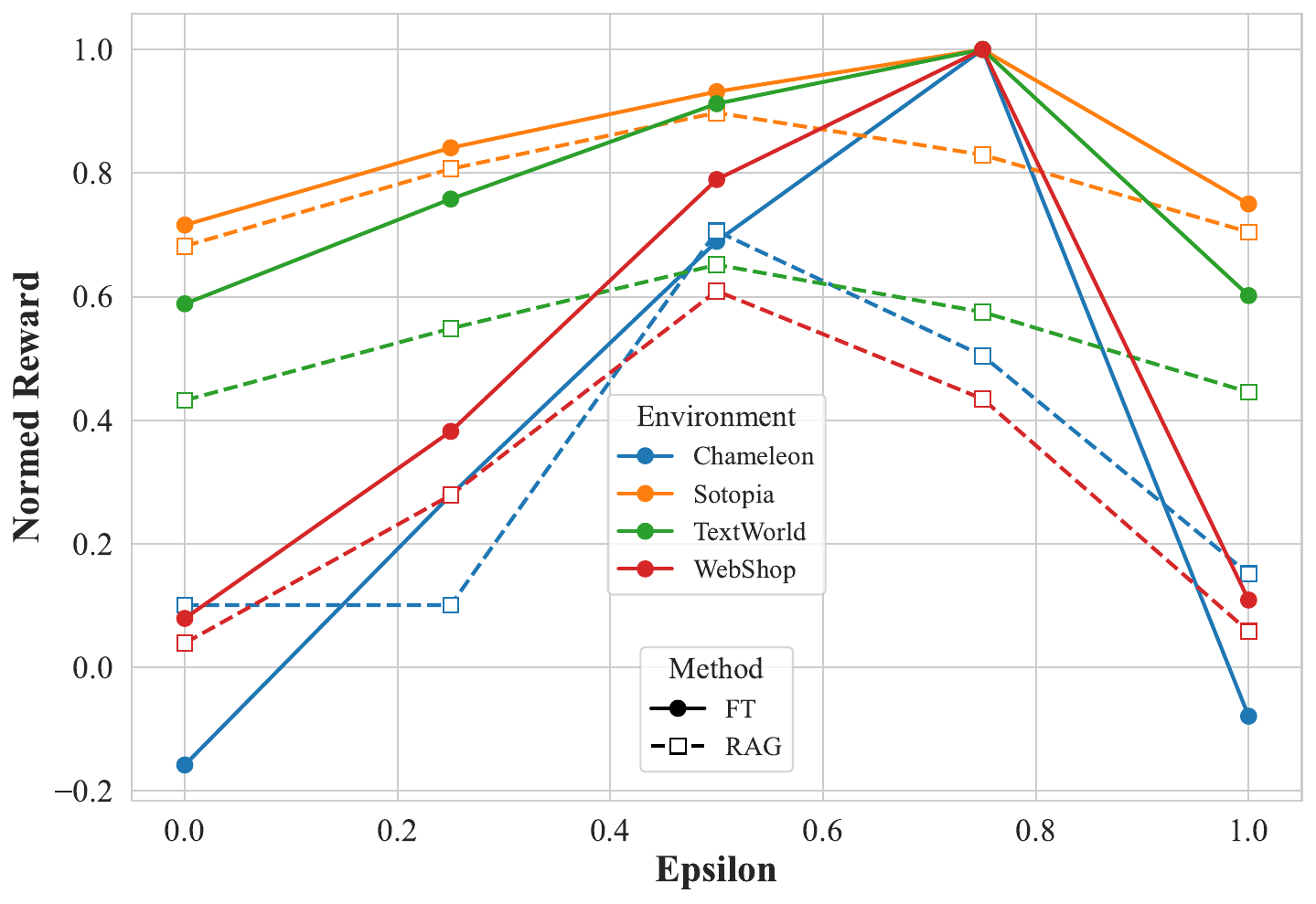}

    \caption{Data scale and exploration $\epsilon$ ablations. RAG is more data-efficient than FT, and both methods benefit from agent exploration.}
    \label{fig:grid_envs}
\end{figure}

\section{Deeper Investigation of RAG-based WMs}

Retrieval-augmented generation (RAG) introduces two distinct points of failure: the retrieval mechanism and the generation process. In standard information retrieval, documents can be evaluated against gold-standard annotations. However, transitions in an experience replay buffer lack pre-annotated relevance labels. Furthermore, a perfectly matching transition may not exist for a given observation-action pair, rendering the quantification of retriever error rates fundamentally ill-posed.

To isolate and evaluate the retrieval step, we introduce a pseudo-relevance scoring mechanism. Assume a world model $\text{WM}(o_t, a_t, e)$ that predicts the next observation given $o_t, a_t$, and a retrieved in-context demonstration $e = (o_i, a_i, o_{i+1})$. Let $\text{diff}(\hat{o}, o_{t+1})$ be an evaluation function (e.g., an LLM-as-a-judge) that quantifies the divergence between the predicted observation $\hat{o}$ and the ground-truth observation $o_{t+1}$. By holding $o_t$ and $a_t$ constant, we counterfactually evaluate candidate transitions from the replay buffer. A transition $e_a$ is considered strictly better than $e_b$ if:
\begin{equation}
    \text{diff}(\text{WM}(o_t, a_t, e_a), o_{t+1}) < \text{diff}(\text{WM}(o_t, a_t, e_b), o_{t+1})
\end{equation}
By applying this function, we can empirically assign a pseudo-relevance score to every transition in the buffer, identifying the oracle transition that explicitly minimises the downstream prediction error. In practice, for each test instance, $(o_t, a_t)$, we compute pseudo-relevance scores only for the top $k=20$ most similar transitions $(e_1\ldots e_k)$ ranked by the cosine similarity between the text embedding of $o_t\circ a_t$ and $o_i\circ a_i, \forall i \in [1, k]$. Our metric of interest is Recall@k ($k\leq5)$, however we note that since we do not compute pseudo-relevance labels over the entire retrieval corpus, the measures reported below are \textbf{upper-bounds} on the recall, i.e., if there is a superior trajectory outside the top-20, it would not affect the recall measure. 
\begin{figure}
    \centering
    \includegraphics[width=0.49\linewidth]{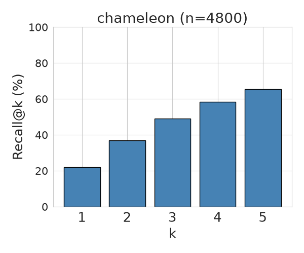}
    \includegraphics[width=0.49\linewidth]{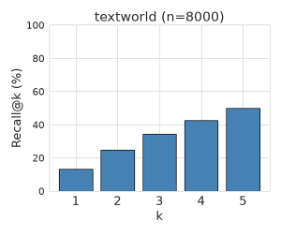}
    \caption{Estimated recall of the retriever on Chameleon and TextWorld. Recall@k is lower than 70\% in both cases, suggesting a substantial gap in performance.}
    \label{fig:retrieval}
\end{figure}

Evaluating dense retrievers against this oracle reveals severely degraded performance (Figure~\ref{fig:retrieval}), with upper bounds on Recall@$5$ plateauing at merely $70\%$. This suboptimal recall indicates that off-the-shelf text embedders, trained primarily for semantic similarity, struggle to capture the dynamics of environment transitions. Consequently, there is headroom for optimisation in the retrieval phase.

To address these limitations, we investigate alternative query formulations. The naive approach concatenates the observation and action into a single query string. However, this forces semantically distinct modalities into a single representational space, frequently causing one feature to overshadow the other. We propose disentangling the representations through two alternative paradigms:

    \noindent \textbf{Linear Combination:} We independently embed the observation and the action. The final retrieval score is a linear combination of their respective cosine similarities. While this successfully preserves individual modalities, its primary limitation is the reliance on a weighting hyperparameter, which is difficult to tune without a representative validation set. In practice, we weight components equally. 
    
    \noindent \textbf{Hierarchical Retrieval:} We structure retrieval as a two-stage cascade. First, we retrieve a broad candidate pool (e.g., top-50) based strictly on observation similarity. Second, we re-rank these candidates based on action similarity. This architecture reflects the intuition that the current state is the primary discriminant for environmental context, while the action serves as a secondary filter within similar states, avoiding the retrieval of identical actions in vastly different environments.

\noindent\textbf{Results:} Evaluating these retrieval paradigms across environments demonstrates that disentangling observation and action representations is generally beneficial, though the method of aggregation is critical. While a linear combination over independent embedding scores improves performance in WebShop (where it serves as the optimal method) and Chameleon, it actively degrades performance below the simple baseline in TextWorld (4.12) and Sotopia (0.514). Hierarchical retrieval proves more robust. The two-stage cascading approach yields the highest total reward in 3 out of 4 environments (TextWorld, Chameleon, and Sotopia). The improvement is particularly stark in TextWorld, where it nearly doubles the reward of the naive baseline (12.62 vs. 6.50). This confirms that prioritising observation similarity before filtering by action is an effective, parameter-free heuristic.

\section{Hybrid World Model}

While retrieval provides dynamic context and fine-tuning embeds domain-specific mechanics, both approaches exhibit isolated limitations. To synergise their strengths, we propose a Hybrid World Model (HWM) that integrates supervised fine-tuning with non-parametric retrieval, explicitly training the language model to condition its predictions on retrieved historical experiences.

To construct the training pipeline, we populate a transition database with historical episodes. For any given target observation-action pair $(o_t, a_t)$, we utilise a dense vector retrieval mechanism to query the database for semantically similar transitions. However, naive retrieval in sequential environments risks severe data contamination. If left unrestricted, a retriever might simply fetch immediately adjacent frames from the same episode, allowing the model to trivially reconstruct the temporal sequence rather than learning underlying environment dynamics.

To prevent this leakage while maximising the utility of the training data, we implement an episode-level masking filter. If a target transition originates from episode $\mathcal{E}_i$, the retriever is constrained to search exclusively within the disjoint set of all available episodes excluding $\mathcal{E}_i$. This strict masking guarantees that the retrieved context acts as a generalised heuristic rather than a direct answer key, preventing intra-episode autocorrelation and ensuring rigorous train-set discipline.

The out-of-episode retrieved transitions are then compiled into a structured prompt alongside the target $o_t$ and $a_t$. We perform supervised fine-tuning on an LM, optimising the standard autoregressive loss on the generation of the ground-truth next observation $o_{t+1}$. By training the model explicitly on retrieval-augmented inputs, the model learns to optimally cross-reference its internal parametric knowledge with the non-parametric context provided by the retrieved demonstrations, yielding a robust and generalizable world model.

\begin{figure}[htbp]
    \centering
    % Row 1: Legend, ALFWorld, TextWorld
    \begin{subfigure}[b]{0.15\textwidth}
        \centering
        \includegraphics[width=\textwidth]{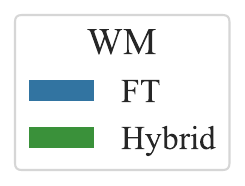}
        \label{fig:ft_rag_legend}
    \end{subfigure}
    \hfill
    \begin{subfigure}[b]{0.32\textwidth}
        \centering
        \includegraphics[width=\textwidth]{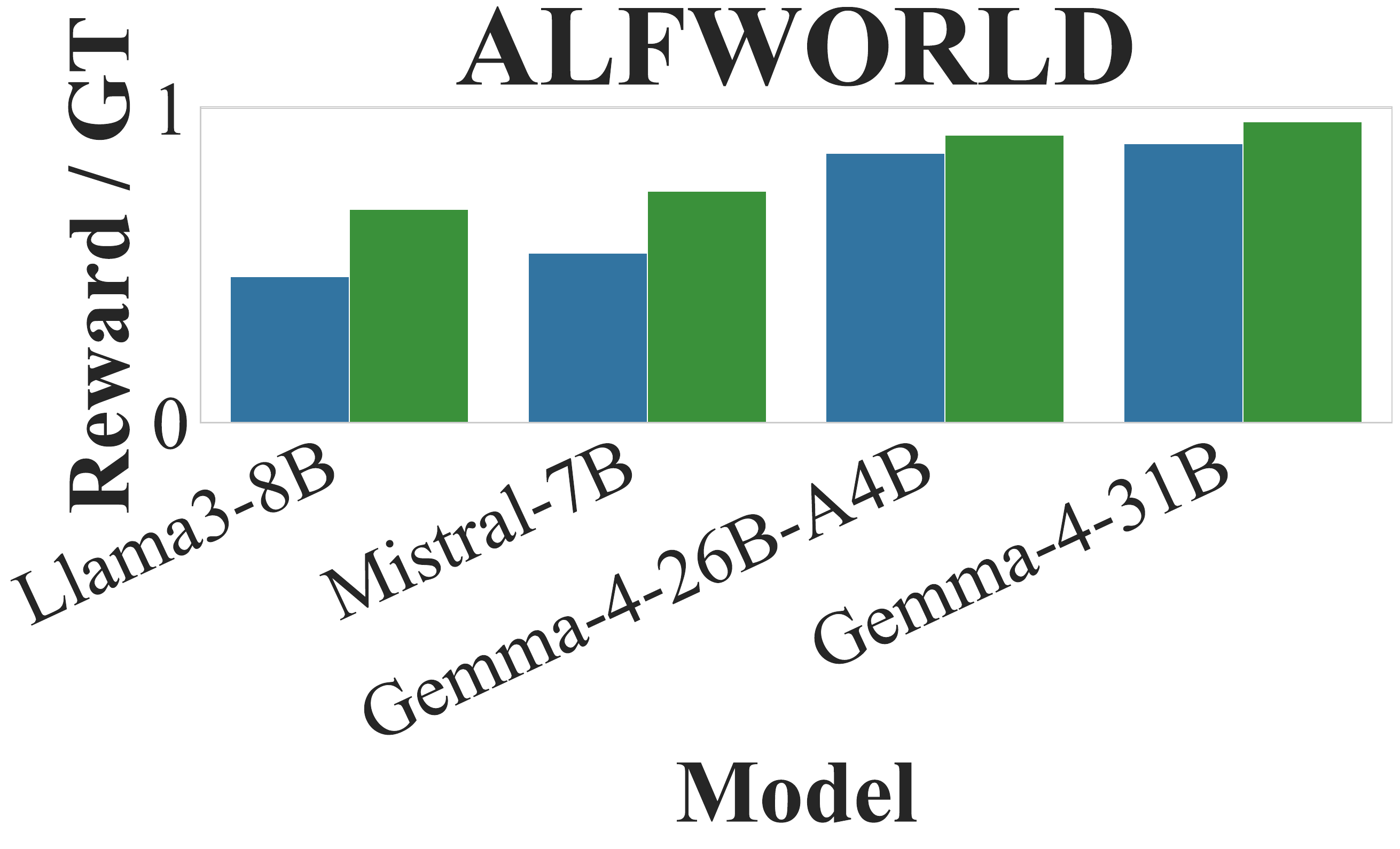}
    \end{subfigure}
    \hfill
    \begin{subfigure}[b]{0.32\textwidth}
        \centering
        \includegraphics[width=\textwidth]{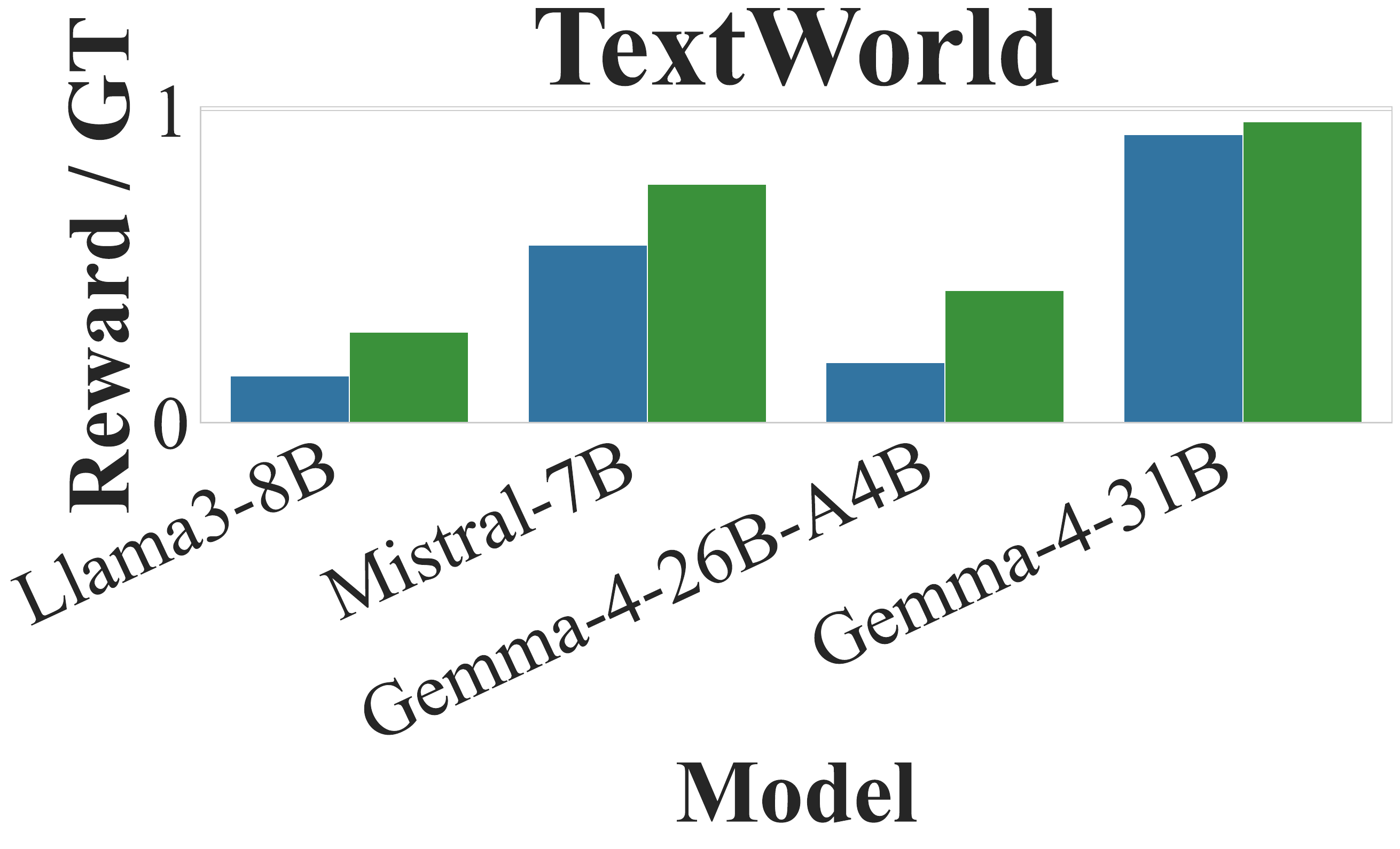}
    \end{subfigure}

    \vspace{1em}

    % Row 2: WebShop, Sotopia, Chameleon
    \begin{subfigure}[b]{0.32\textwidth}
        \centering
        \includegraphics[width=\textwidth]{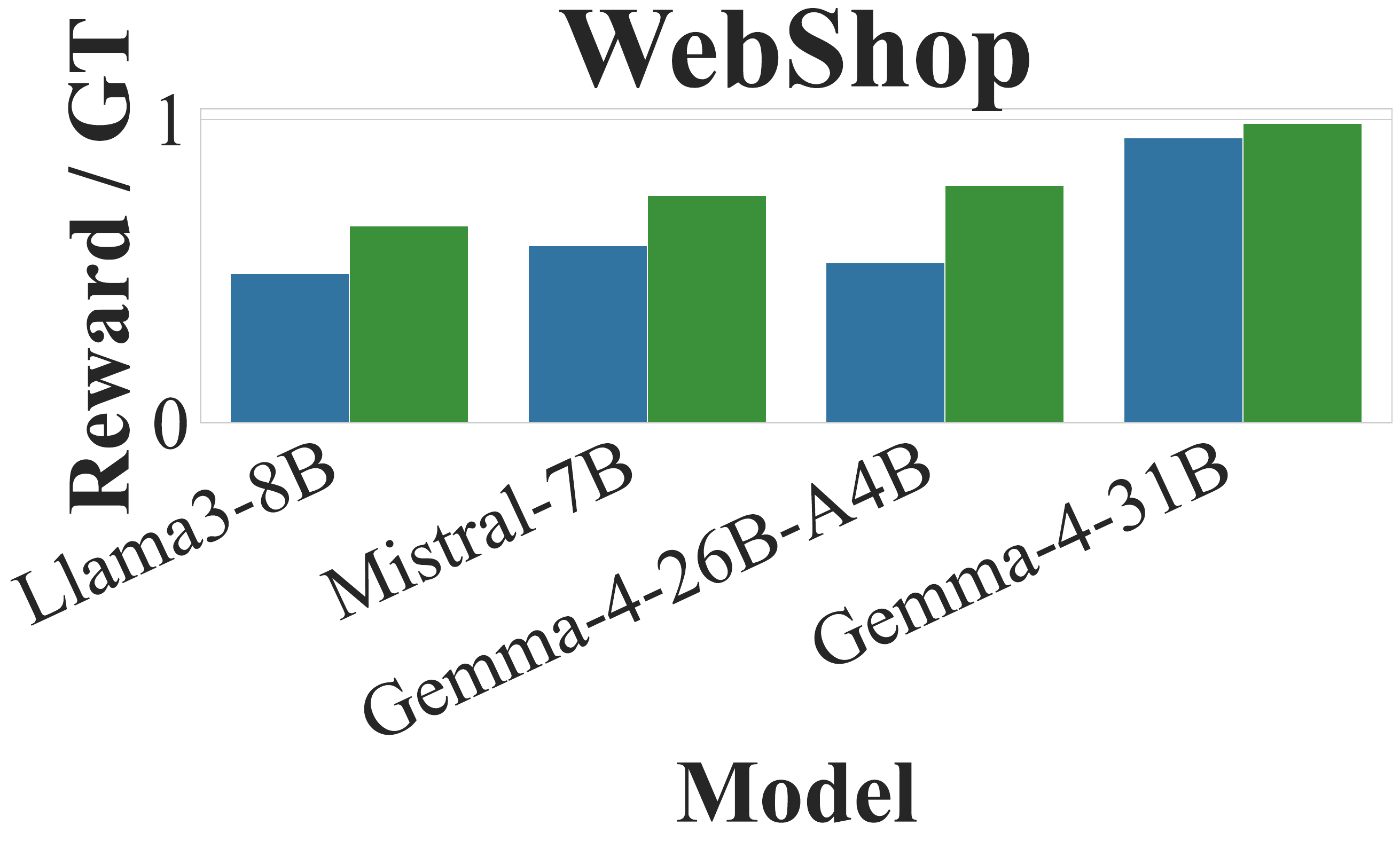}
    \end{subfigure}
    \hfill
    \begin{subfigure}[b]{0.32\textwidth}
        \centering
        \includegraphics[width=\textwidth]{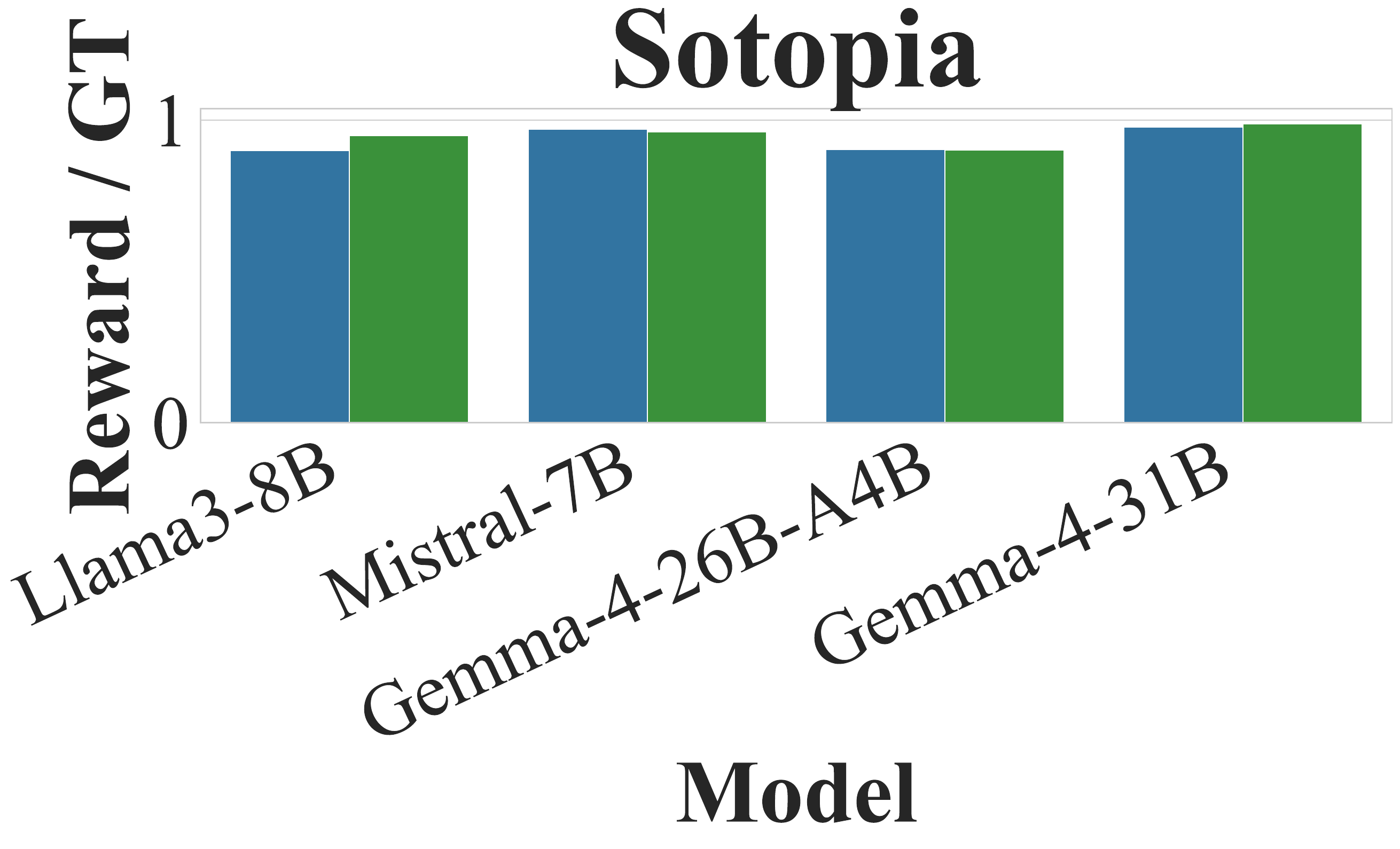}
    \end{subfigure}
    \hfill
    \begin{subfigure}[b]{0.32\textwidth}
        \centering
        \includegraphics[width=\textwidth]{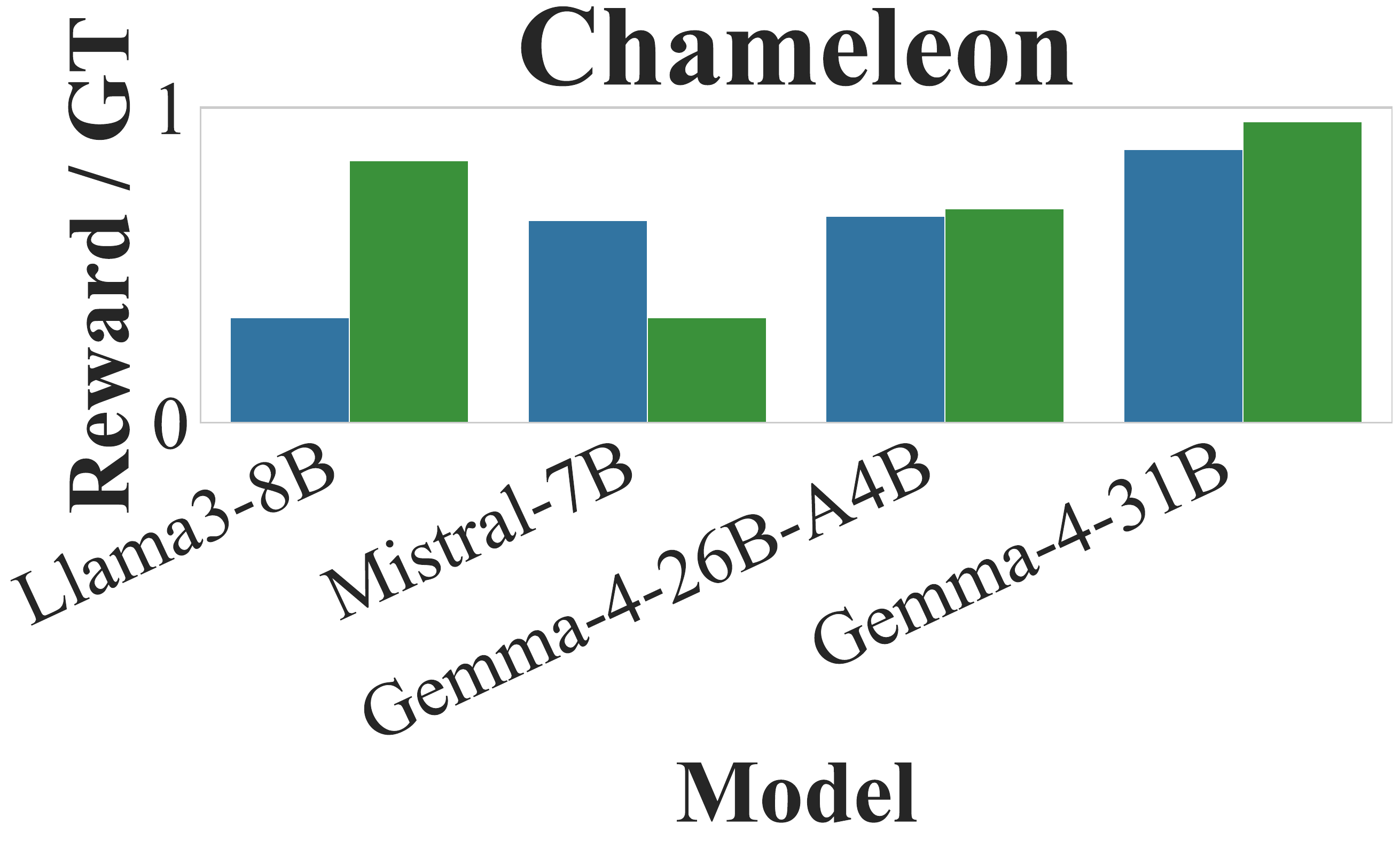}
    \end{subfigure}

    \caption{Our hybrid approach combines the benefits of fine-tuning and RAG, outperforming other approaches on 17/20 settings at $p<0.05$. For the complete results with CIs, see Appendix~\ref{sec:appcomplete}.}
    \label{fig:ft_rag_grid}
\end{figure}

\noindent\textbf{Results:} Our hybrid approach outperforms on 17/20 settings, resolving the bottlenecks of isolated fine-tuning and naive retrieval. The method reaches 77\% of the total reward obtainable by the oracle, substantially outperforming the next-best baseline (60\%). Furthermore, on 45\% of all environment-model combinations, the hybrid model exceeds 90\% of oracle performance, showing that fusing internal weight updates with non-parametric context yields highly robust world models.

\section{Conclusion}
\label{sec:conclusion}

In this work, we perform the first systematic comparison of RAG and fine-tuning for the task of LM-based world modelling. We show that unlike tasks like knowledge acquisition, fine-tuning proves consistently superior to RAG, with only minor exceptions. We demonstrate that while both approaches benefit from increased exploration, RAG is more data-efficient. Our error analysis reveals substantial retriever errors, suggesting that adapting the retriever to the specific environment could yield substantial performance gains. We investigate alternative paradigms of encoding transitions and show that a hierarchical scheme outperforms the classic encoding method. Finally, we introduce a hybrid world model system that combines fine-tuning and RAG to consistently outperform all other baselines. 

\bibliography{iclr2027_conference}
\bibliographystyle{iclr2027_conference}

\appendix

\newpage

\section{Environment Implementation Details}
\label{app:env-details}

Across all five environments, the world model (WM) is trained to predict the agent's next observation given the interaction history and a candidate action; unless otherwise noted below, the training target is simply the environment's true next observation.

\subsection{Embodied Navigation: Observation Shaping}

\paragraph{ALFWorld.} Rather than exposing the raw simulator observation on its own, each observation given to the agent is augmented with the current room description, the current inventory, and the full list of admissible commands, so that the agent always has a complete, self-contained view of the state at every step. On reset, the environment's introductory banner text is discarded and the task instruction is extracted and surfaced separately from the observation, so that the very first observation of an episode has the same shape as every subsequent one. No truncation is applied to any of these fields.

\paragraph{TextWorld.} The same augmentation strategy is used (room description, inventory, admissible commands appended to the observation), with one deliberate deviation: the \emph{look} command is removed from the admissible-command list before it is shown to the agent, since taking that action would simply reproduce the room description already present in the observation and offers no new information. As with ALFWorld, introductory/banner text is dropped at reset and replaced with the task objective, so reset and step observations share a consistent format.

\paragraph{WebShop.} Observations are shaped in three ways before being shown to the agent. First, the shopping instruction — which the underlying environment otherwise repeats verbatim inside every observation — is removed, since it is instead surfaced once as a fixed task description for the episode. Second, on the terminal observation, the trailing reward-summary text that the environment appends is stripped, since reward is already tracked numerically by the wrapper rather than needing to be parsed out of the observation text. Third, the set of currently available actions (whether a search is possible, and which clickable page elements are currently valid) is appended explicitly to the observation as an action menu, so the agent's action space at each step is stated rather than implicit. No additional truncation or markup handling is applied; the underlying text rendering of search-results and product pages is otherwise used as-is.

\subsection{Adversarial Negotiation: Sotopia}

\paragraph{Setup.} Each episode is a two-party conversation grounded in a scenario and a pair of character profiles (physical/social context, occupation, personality, private goal, and a piece of secret information each party is not expected to reveal unless it serves them), drawn from the public Sotopia scenario dataset. The agent under evaluation always plays one fixed conversational role; the other party is a separate, non-evaluated language model instantiated specifically to play the counterpart for that episode.

\paragraph{Prompting.} Each turn, the speaking party is prompted with the scenario description, its own persona (name, age, occupation, publicly known information, personality and values), its private goal, its secret information, and the conversation so far, and is asked to produce the single best next utterance to advance its own goal, or to end the interaction if it has reached a natural conclusion.

\paragraph{Action and observation space.} The action space is an unconstrained natural-language utterance, or an explicit "end the conversation" action available to either party. After each of the agent's turns, the observation returned is the counterpart's reply since the agent's last turn; the agent's own utterance is not echoed back, since it is already recorded as the action that produced that step.

\paragraph{Termination and turn budget.} An episode ends when either party chooses to end the conversation, or after a fixed budget of ten player turns has elapsed — chosen to allow conversations to develop fully before being cut off.

\paragraph{Reward.} At the end of an episode, a separate judging language model rates the evaluated party's goal achievement on a 0--10 scale, conditioned on the scenario, that party's persona and goal, and the full conversation transcript; this score is rescaled to $[0,1]$ to form the scalar reward.

\paragraph{Data splits.} Scenarios are partitioned by their labeled interaction type. Negotiation scenarios are split 80/20 by scenario into train and test sets under a fixed random seed. Scenarios labeled as competitive (rather than negotiation) are held out in their entirety and used as a distribution-shift evaluation split, disjoint from both the negotiation train and test scenarios.

\paragraph{World-model prediction task.} Given the conversation history, the current observation, and the agent's proposed next utterance as the candidate action, the WM is trained to predict the counterpart's resulting reply — i.e.\ the ground-truth prediction target is literally the counterpart's actual next line of dialogue (or the episode's terminal state, if the proposed utterance ends the conversation).

\subsection{Deceptive Communication: Chameleon}

\paragraph{Default configuration.} The Chameleon environment is general enough to support an arbitrary number of players and an arbitrary number of chameleons among them, but every experiment reported in this paper uses a single fixed configuration: three players and exactly one chameleon. The evaluated agent always occupies the first speaking position and is always a non-chameleon; the chameleon role is always assigned to the second speaking position, leaving the third position as the other (non-chameleon, language-model-driven) player. Three players is in any case the smallest group size for which the game is well defined, and with three players exactly one chameleon is the only valid configuration, so this default is also the game's minimal, canonical form.

\paragraph{Game flow.} Each round is drawn from one of several topics (e.g.\ movies, athletes, companies, countries, superheroes, scientists, food), each with a fixed pool of sixteen candidate secret words per data split. Players speak once each, in a fixed order. Non-chameleon players are shown the secret word and asked to give a single related word that signals their knowledge of it to other non-chameleons without making it obvious to the chameleon; the chameleon does not know the secret word and must produce a plausible-sounding clue by inference from the other clues and the topic alone. Since the evaluated agent always speaks first, it must commit to its clue with no information beyond the topic and secret word, and the round always concludes after everyone has spoken once.

\paragraph{Judge.} After every player has given a clue, a separate judging language model is shown the secret word, the full candidate word list for the topic, and all clues in speaking order, and must name exactly one player as the chameleon.

\paragraph{Reward.} From the evaluated player's perspective (always a non-chameleon under the default configuration): the reward is $+1$ if the judge correctly names the true chameleon, $0$ if the judge instead accuses the other non-chameleon player, and this outcome would additionally be penalized by $-1$ if the judge accused the evaluated player itself — a case that cannot arise under the fixed default configuration, since the evaluated player's position is never assigned the chameleon role.

\paragraph{World-model prediction task.} Because the round always ends after a single exchange of clues, the world model's prediction task is correspondingly single-step: conditioned on the topic, secret word, and the evaluated agent's chosen clue (the candidate action), the WM must predict both the clues subsequently given by the other two players and the judge's eventual verdict. Rather than using the literal next observation as the training target, the ground-truth target is constructed specifically to isolate these two pieces of information: the other players' clues, and a short natural-language statement of whether the judge's guess correctly caught the true chameleon or was misled into accusing the wrong player.

\section{World Model Deployment and Evaluation}
\label{app:wm-deployment}

At deployment time, the policy-agent LM is prompted with the environment description, task description, interaction history, and current observation, and asked to produce a ranked list of $m$ candidate actions in a single call, ordered from most to least preferred, rather than a single committed action. The base framing shared by every prompt to the policy-agent LM is:

\begin{tcolorbox}[breakable]
\ttfamily
You are an agent in [ENVIRONMENT DESCRIPTION]\\
Your task is: [TASK\_DESCRIPTION]

The history of your interactions with the environment is:\\{}
[HISTORY]

Your current observation of the environment is:\\{}
[OBS]
\end{tcolorbox}

\noindent and the specific instruction appended to it to elicit the ranked candidate list is:

\begin{tcolorbox}[breakable]
\ttfamily
Given this, what are the best actions to take next? Give a ranked list of the top [K] action candidates, in order of preference, with the most preferred action first. Give your output in the following format:

Reasoning: [your reasoning here, which will not be included in the final output]\\
Action 1: [the most preferred action, exactly as it should be input into the environment, and nothing else]\\
Action 2: [the second most preferred action, exactly as it should be input into the environment, and nothing else]\\
...\\
Action K: [the K-th most preferred action, exactly as it should be input into the environment, and nothing else]\\
{[}STOP{]}
\end{tcolorbox}

Each of the $m$ candidates is then passed to the world model under evaluation, which independently predicts the resulting next observation for that candidate given the same history and current observation. The prompt used to elicit each prediction is:

\begin{tcolorbox}[breakable]
\ttfamily
You are a world model in [ENVIRONMENT DESCRIPTION]

[RETRIEVED TRANSITION STRING]

The history of the agent's interactions with the environment is:\\{}
[HISTORY]

The current observation of the environment is:\\{}
[OBS]

The agent is considering taking this action:\\{}
[ACTION]

Given all of this what is the most likely change in state or next observation that the agent will see?\\
First reason step by step about the mechanics, then describe the resulting state change.\\
Give your answer in the following format:\\
Reasoning: [your reasoning here, which will not be included in the final output]\\
Prediction: [a concise description of what changed and what the new state of the environment looks like]\\
{[}STOP{]}
\end{tcolorbox}

\noindent (the retrieved-transition block is populated only for retrieval-augmented world models, and is left empty otherwise.) This yields $m$ predicted observations, one per candidate action.

Each predicted observation is then scored by the same policy-agent LM in a separate call, using the same base framing above, followed by:

\begin{tcolorbox}[breakable]
\ttfamily
You then took action [ACTION] and a state change occurred in the environment, giving you the next observation:\\{}
[STATE\_CHANGE]

Given this, how beneficial is this state change for achieving your task? Rate on a scale of 1 to 10, where 1 means active regression from the goal, 3 means no real progress, 5 means some progress, 7 means significant progress, and 10 means excellent progress or goal achieved.\\
First reason step by step about how this state change relates to your task, then give your final score.\\
Answer in the following format:\\
Reasoning: \textless think about how this state change relates to the task goal\textgreater\\
Score: \textless integer 1 to 10\textgreater\\
{[}STOP{]}
\end{tcolorbox}

The candidate action whose predicted observation receives the highest score is selected and executed in the real environment. If the world model fails to produce a usable prediction for any of the $m$ candidates, the agent falls back to executing its own top-ranked candidate directly, without invoking the scoring step.

In practice we set $m=5$. The policy-agent LM is held frozen throughout — it is never fine-tuned or otherwise updated over the course of these experiments — and is used identically for candidate-action proposal and for scoring predicted observations. We evaluate this pipeline with four different policy-agent LMs: Llama-3.1-8B-Instruct~\citep{grattafiori2024llama}, Mistral-7B-Instruct~\citep{Jiang2023Mistral7}, Gemma-4-26B-A4B-Instruct, and Gemma-4-31B-Instruct~\citep{team2025gemma}.

\section{Exploration}
\label{app:exploration}

Training buffers for the world models are constructed by rolling out an agent in each training environment. At each step, with probability $\epsilon$ the agent's action is instead generated by an exploratory variant of the same policy-agent LM rather than its default goal-directed behavior; with probability $1-\epsilon$ it behaves greedily as usual. In both cases the action is still produced by prompting the LM with the same base framing and output format shown in Appendix~\ref{app:wm-deployment} (single action or ranked candidate list, depending on whether a world model is active that step) — exploration is implemented entirely by substituting the task instruction given to the model, not by sampling actions uniformly at random.

For most environments, the exploratory variant entirely replaces \texttt{[TASK\_DESCRIPTION]} in the base framing above with a single generic exploration instruction:

\begin{tcolorbox}[breakable]
\ttfamily
Take interesting exploratory actions to probe various aspects of the environment. Avoid doing the same thing repeatedly, and try to think creatively and be very wide in what you consider doing (but within any specified list of allowed actions).
\end{tcolorbox}

Two environments instead use a specialized exploration instruction tailored to their setting, appended to (rather than replacing) the normal task instruction, in the form ``\ttfamily [normal task instruction]\rmfamily\ [line break] For this turn, try an exploratory approach: \ttfamily [specialized instruction] \rmfamily''. In Chameleon, the specialized instruction is:

\begin{tcolorbox}[breakable]
\ttfamily
Be radically creative with your clue. Ignore what seems safe or obvious — pick something that no one would expect, that comes from a completely different angle, that might even seem wrong at first glance but is secretly defensible. The goal is maximum surprise while still being technically valid. Push hard against your first instinct.
\end{tcolorbox}

\noindent and in Sotopia, it is:

\begin{tcolorbox}[breakable]
\ttfamily
Throw out the safe playbook entirely. Take a bold, unexpected gamble — try something you normally wouldn't dare, a completely different angle, an unconventional tactic that could pay off big or fail spectacularly. The worst outcome is being boring. Commit fully to the unusual choice.
\end{tcolorbox}

This exploration mechanism is applied uniformly across all environments and is independent of which world model, if any, is being used to select actions during rollout. It is used only when constructing training buffers; at evaluation time $\epsilon$ is fixed to zero, so that reported results always reflect purely greedy behavior. We sweep $\epsilon \in \{0, 0.25, 0.5, 0.75, 1.0\}$ when constructing training buffers, in order to study how the mixture of exploratory and greedy trajectories in the training buffer affects the resulting world model's quality.

\section{Complete Results}
\label{sec:appcomplete}

We show complete results, including 95\% confidence intervals from a bootstrap t-test below (Table~\ref{table:corepm}). We also show the fidelity measurements for the Llama3-8B model on all explored metrics (Table~\ref{table:fullapp}). Limited variance for BLEU and BertScore suggest that they are poor metrics for WM fidelity. 

% Please add the following required packages to your document preamble:
% \usepackage{booktabs}
% \usepackage[table,xcdraw]{xcolor}
% \usepackage{graphicx}
% Beamer presentation requires \usepackage{colortbl} instead of \usepackage[table,xcdraw]{xcolor}
\begin{sidewaystable}[]
\centering
\footnotesize
\setlength{\tabcolsep}{4pt}
\resizebox{\linewidth}{!}{%
\begin{tabular}{@{}lllllllllllll@{}}
\toprule
Environment & Model & No & \multicolumn{2}{c}{ZS} & \multicolumn{2}{c}{FS} & \multicolumn{2}{c}{FT} & \multicolumn{2}{c}{RAG} & \multicolumn{2}{c}{Hybrid} \\ \midrule
 & & Util & Util & Fid & Util & Fid & Util & Fid & Util & Fid & Util & Fid \\ \midrule
ALF          & Llama    & $0.36 \pm 0.02$          & $0.43 \pm 0.01$          & $4.15 \pm 0.02$          & $0.25 \pm 0.06$          & $3.96 \pm 0.08$          & $\underline{0.46} \pm 0.01$ & $\underline{4.78} \pm 0.19$ & $0.36 \pm 0.00$          & $4.09 \pm 0.02$          & $\mathbf{0.68} \pm 0.06$ & $\mathbf{7.42} \pm 0.79$ \\
             & Mistral  & $0.00 \pm 0.12$          & $0.30 \pm 0.03$          & $3.95 \pm 0.05$          & $0.20 \pm 0.06$          & $3.82 \pm 0.08$          & $\underline{0.54} \pm 0.05$ & $\underline{4.65} \pm 0.16$ & $0.39 \pm 0.03$          & $4.12 \pm 0.05$          & $\mathbf{0.74} \pm 0.06$ & $\mathbf{7.55} \pm 0.87$ \\
             & Gem-26B  & $0.79 \pm 0.01$          & $0.80 \pm 0.00$          & $4.33 \pm 0.16$          & $0.76 \pm 0.01$          & $4.16 \pm 0.10$          & $\underline{0.86} \pm 0.01$ & $\underline{5.10} \pm 0.07$ & $0.81 \pm 0.00$          & $4.87 \pm 0.07$          & $\mathbf{0.92} \pm 0.02$ & $\mathbf{6.12} \pm 0.31$ \\
             & Gem-31B  & $0.82 \pm 0.01$          & $0.67 \pm 0.03$          & $4.23 \pm 0.47$          & $0.72 \pm 0.06$          & $5.01 \pm 0.03$          & $\underline{0.89} \pm 0.01$ & $\underline{6.18} \pm 0.32$ & $0.86 \pm 0.01$          & $5.11 \pm 0.03$          & $\mathbf{0.96} \pm 0.02$ & $\mathbf{7.84} \pm 0.50$ \\ \midrule
Text         & Llama    & $0.00 \pm 0.04$          & $\underline{0.16} \pm 0.00$ & $3.40 \pm 0.19$          & $0.06 \pm 0.03$          & $2.87 \pm 0.32$          & $0.15 \pm 0.00$          & $4.04 \pm 0.03$          & $0.11 \pm 0.01$          & $\underline{4.15} \pm 0.03$ & $\mathbf{0.29} \pm 0.04$ & $\mathbf{4.42} \pm 0.08$ \\
             & Mistral  & $0.494 \pm 0.006$        & $\underline{0.634} \pm 0.020$ & $3.45 \pm 0.19$          & $0.504 \pm 0.028$        & $2.90 \pm 0.33$          & $0.569 \pm 0.005$        & $4.10 \pm 0.03$          & $0.552 \pm 0.005$        & $\underline{4.20} \pm 0.03$ & $\mathbf{0.764} \pm 0.039$ & $\mathbf{4.55} \pm 0.10$ \\
             & Gem-26B  & $0.000 \pm 0.087$        & $\underline{0.219} \pm 0.008$ & $3.61 \pm 0.20$          & $0.144 \pm 0.014$        & $3.13 \pm 0.29$          & $0.193 \pm 0.007$        & $4.28 \pm 0.02$          & $0.168 \pm 0.007$        & $\underline{4.36} \pm 0.02$ & $\mathbf{0.424} \pm 0.061$ & $\mathbf{5.17} \pm 0.24$ \\
             & Gem-31B  & $0.848 \pm 0.017$        & $0.755 \pm 0.055$        & $5.20 \pm 0.13$          & $0.631 \pm 0.075$        & $4.82 \pm 0.23$          & $\underline{0.924} \pm 0.006$ & $\underline{5.86} \pm 0.07$ & $0.904 \pm 0.006$        & $5.63 \pm 0.07$          & $\mathbf{0.964} \pm 0.012$ & $\mathbf{6.84} \pm 0.29$ \\ \midrule
Web          & Llama    & $0.04 \pm 0.02$          & $0.07 \pm 0.01$          & $4.77 \pm 0.09$          & $0.09 \pm 0.05$          & $3.78 \pm 0.59$          & $\underline{0.49} \pm 0.05$ & $\underline{6.06} \pm 0.23$ & $0.24 \pm 0.05$          & $5.08 \pm 0.09$          & $\mathbf{0.65} \pm 0.05$ & $\mathbf{6.84} \pm 0.23$ \\
             & Mistral  & $0.00 \pm 0.10$          & $0.17 \pm 0.05$          & $4.65 \pm 0.15$          & $0.25 \pm 0.02$          & $3.85 \pm 0.48$          & $\underline{0.58} \pm 0.05$ & $\underline{6.15} \pm 0.24$ & $0.33 \pm 0.02$          & $5.15 \pm 0.15$          & $\mathbf{0.75} \pm 0.05$ & $\mathbf{6.95} \pm 0.24$ \\
             & Gem-26B  & $0.08 \pm 0.02$          & $0.12 \pm 0.03$          & $4.81 \pm 0.16$          & $0.17 \pm 0.05$          & $4.11 \pm 0.42$          & $\underline{0.53} \pm 0.05$ & $\underline{6.27} \pm 0.19$ & $0.35 \pm 0.05$          & $5.34 \pm 0.16$          & $\mathbf{0.78} \pm 0.08$ & $\mathbf{6.89} \pm 0.19$ \\
             & Gem-31B  & $0.904 \pm 0.006$        & $0.601 \pm 0.052$        & $4.01 \pm 0.17$          & $0.687 \pm 0.130$        & $4.29 \pm 0.46$          & $\underline{0.940} \pm 0.005$ & $\underline{6.43} \pm 0.05$ & $0.923 \pm 0.005$        & $5.81 \pm 0.19$          & $\mathbf{0.987} \pm 0.014$ & $\mathbf{6.58} \pm 0.05$ \\ \midrule
Cha          & Llama    & $0.17 \pm 0.05$          & $-0.50 \pm 0.24$         & $2.56 \pm 0.08$          & $-0.10 \pm 0.08$         & $2.70 \pm 0.25$          & $-0.52 \pm 0.01$         & $3.54 \pm 0.10$          & $\underline{0.33} \pm 0.05$ & $\underline{3.87} \pm 0.10$ & $\mathbf{0.83} \pm 0.15$ & $\mathbf{4.88} \pm 0.30$ \\
             & Mistral  & $-4.50 \pm 0.37$         & $-3.74 \pm 0.20$         & $2.45 \pm 0.09$          & $-3.07 \pm 0.20$         & $2.60 \pm 0.26$          & $-3.88 \pm 0.09$         & $3.45 \pm 0.15$          & $\mathbf{0.64} \pm 0.09$ & $\underline{3.95} \pm 0.05$ & $\underline{0.33} \pm 0.09$ & $\mathbf{4.13} \pm 0.05$ \\
             & Gem-26B  & $0.501 \pm 0.006$        & $0.070 \pm 0.246$        & $3.22 \pm 0.12$          & $0.480 \pm 0.006$        & $3.42 \pm 0.08$          & $0.480 \pm 0.000$        & $3.70 \pm 0.08$          & $\underline{0.655} \pm 0.007$ & $\underline{4.31} \pm 0.11$ & $\mathbf{0.679} \pm 0.007$ & $\mathbf{4.66} \pm 0.11$ \\
             & Gem-31B  & $0.78 \pm 0.01$          & $-0.05 \pm 0.27$         & $3.28 \pm 0.33$          & $0.41 \pm 0.20$          & $4.16 \pm 0.10$          & $0.73 \pm 0.01$          & $3.83 \pm 0.10$          & $\underline{0.87} \pm 0.03$ & $\underline{5.17} \pm 0.30$ & $\mathbf{0.96} \pm 0.03$ & $\mathbf{6.24} \pm 0.32$ \\ \midrule
Soto         & Llama    & $0.75 \pm 0.04$          & $0.89 \pm 0.00$          & $4.50 \pm 0.03$          & $0.82 \pm 0.02$          & $4.19 \pm 0.19$          & $\underline{0.90} \pm 0.00$ & $4.61 \pm 0.03$          & $0.86 \pm 0.01$          & $\underline{4.77} \pm 0.04$ & $\mathbf{0.95} \pm 0.01$ & $\mathbf{4.90} \pm 0.04$ \\
             & Mistral  & $0.950 \pm 0.004$        & $0.963 \pm 0.000$        & $4.40 \pm 0.05$          & $0.956 \pm 0.004$        & $4.24 \pm 0.10$          & $\mathbf{0.970} \pm 0.000$ & $4.58 \pm 0.05$          & $\underline{0.969} \pm 0.000$ & $\underline{4.86} \pm 0.02$ & $0.961 \pm 0.000$        & $\mathbf{4.92} \pm 0.02$ \\
             & Gem-26B  & $0.850 \pm 0.010$        & $0.868 \pm 0.001$        & $4.61 \pm 0.09$          & $0.869 \pm 0.006$        & $4.42 \pm 0.11$          & $\mathbf{0.903} \pm 0.000$ & $4.92 \pm 0.05$          & $0.887 \pm 0.004$        & $\mathbf{5.12} \pm 0.01$ & $\underline{0.902} \pm 0.000$ & $\underline{5.07} \pm 0.01$ \\
             & Gem-31B  & $0.96 \pm 0.00$          & $0.96 \pm 0.00$          & $4.73 \pm 0.02$          & $0.64 \pm 0.19$          & $4.62 \pm 0.07$          & $\underline{0.98} \pm 0.00$ & $4.81 \pm 0.02$          & $0.97 \pm 0.00$          & $\underline{4.87} \pm 0.01$ & $\mathbf{0.99} \pm 0.00$ & $\mathbf{4.92} \pm 0.01$ \\ \bottomrule
\end{tabular}
}
\caption{Utility and Fidelity of LM-based WMs. \textbf{Bold} marks the best value in each row per metric; \underline{underline} marks the second best. Fine-tuning is the superior standalone method, achieving superior utility on 15/20 settings. RAG maintains a consistent advantage over the baseline approaches and outperforms fine-tuning on the Chameleon environment. Hybrid outperforms all other methods consistently, with the only exceptions being a single setting of Chameleon and two Sotopia settings.}
\label{table:corepm}
\end{sidewaystable}

\begin{table}[]
\centering
\begin{tabular}{@{}llrrr@{}}
\toprule
Environment & WM & BLEU & BertScore & LM-as-Judge \\ \midrule
ALFWORLD & None &  &  & - \\
 & ZS & 0.00 & 0.83 & 4.15 \\
 & FS & 0.00 & 0.79 & 3.96 \\
 & FT & 0.00 & 0.80 & 4.78 \\
 & RAG & 0.00 & 0.89 & 4.09 \\
 & Hybrid & 0.02 & 0.85 & 7.42 \\
 & GT &  &  & 10.00 \\ \midrule
Chameleon & None & 0.00 &  & - \\
 & ZS & 0.00 & 0.81 & 2.56 \\
 & FS & 0.02 & 0.83 & 2.70 \\
 & FT & 0.04 & 0.83 & 3.54 \\
 & RAG & 0.02 & 0.85 & 3.87 \\
 & Hybrid & 0.03 & 0.84 & 4.88 \\
 & GT &  &  & 10.00 \\ \midrule
Sotopia & None &  &  & - \\
 & ZS & 0.03 & 0.84 & 4.50 \\
 & FS & 0.03 & 0.85 & 4.19 \\
 & FT & 0.03 & 0.85 & 4.61 \\
 & RAG & 0.18 & 0.91 & 4.77 \\
 & Hybrid & 0.21 & 0.90 & 4.90 \\
 & GT &  &  & 10.00 \\ \midrule
TextWorld & None &  &  & - \\
 & ZS & 0.01 & 0.76 & 3.40 \\
 & FS & 0.01 & 0.76 & 2.87 \\
 & FT & 0.01 & 0.76 & 4.04 \\
 & RAG & 0.00 & 0.78 & 4.15 \\
 & Hybrid & 0.04 & 0.77 & 4.42 \\
 & GT &  &  & 10.00 \\ \midrule
WebShop & None &  &  & - \\
 & ZS & 0.01 & 0.78 & 4.77 \\
 & FS & 0.00 & 0.78 & 3.78 \\
 & FT & 0.01 & 0.77 & 6.06 \\
 & RAG & 0.01 & 0.78 & 5.08 \\
 & Hybrid & 0.30 & 0.90 & 6.84 \\
 & GT &  &  & 10.00 \\ \bottomrule
\end{tabular}
\caption{Complete results with all fidelity metrics. Low variance with BLEU and BertScore suggest they are poor metrics for WM-fidelity.}
\label{table:fullapp}
\end{table}

\section{Hardware}
Experiments were run on a single cluster with 8xA100 NVIDIA GPUs. 

\section{Use of LLMs}
We used LLMs for the following:
\begin{itemize}
    \item Help with code implementation (Claude Code). No design decisions were made unilaterally by the LLM. 
    \item Appendix writing: Claude Code was used to verbalize the experimental details and hyperparameters during the writing of certain Appendix sections. All text was checked by the authors to confirm veracity.
    \item Error and spell checking. 
\end{itemize}

We did not use LLMs for any of the following:
\begin{itemize}
    \item Ideation/project design
    \item Interpretation of results
\end{itemize}

\end{document}